\documentclass{article}

\PassOptionsToPackage{numbers, compress}{natbib}

\usepackage[preprint]{neurips_2024}

\usepackage{amsmath}
\usepackage[utf8]{inputenc} 
\usepackage[T1]{fontenc}    
\usepackage{hyperref}       
\usepackage{url}            
\usepackage{booktabs}       
\usepackage{amsfonts}       
\usepackage{nicefrac}       
\usepackage{microtype}      
\usepackage{xcolor}         
\usepackage{graphicx}
\usepackage{caption}
\usepackage{changepage}

\usepackage{color}
\definecolor{citecolor}{RGB}{66,168,235}
\definecolor{linkcolor}{RGB}{255,0,0}
\hypersetup{colorlinks=true,citecolor=citecolor,linkcolor=linkcolor}

\usepackage{enumitem}
\usepackage{wrapfig}

\title{FreeTuner: Any Subject in  Any Style with Training-free Diffusion
}

\author{%
  Youcan Xu$^{1}$\thanks{Youcan and Zhen are co-first authors with equal contributions, $^{\dagger}$Long is the corresponding author.}, \; Zhen Wang$^{1,2*}$, \; Jun Xiao$^1$, \; Wei Liu$^3$, \; Long Chen$^{2\dagger}$ \\
  $^1$Zhejiang University \; $^2$Hong Kong University of Science and Technology \; $^3$Tencent \\
  \texttt{youcanxv@163.com; zju\_wangzhen@zju.edu.cn; longchen@ust.hk} \\
}

\begin{document}
\maketitle
\newcommand{\name}{BLIP-Diffusion}
\def\DX#1{{\color{red}{\bf [DX:}{\it{#1}}{\bf ]}}}
\newcommand{\ie}{\emph{i.e.}}
\newcommand{\eg}{\emph{e.g.}}
\newcommand{\Fig}{Figure}
\newcommand{\cf}{\emph{c.f.}}
\newcommand{\etc}{etc}
\newcommand{\lc}[1]{{\color{blue}{$^\textbf{\emph{Long:}}$[#1]}}}

\newcommand{\wz}[1]{{\color{blue}{$^\textbf{\emph{Zhen:}}$[#1]}}}

\begin{figure*}[!h]
\centering
    \vspace{-1em}
    \includegraphics[width=0.95\textwidth]{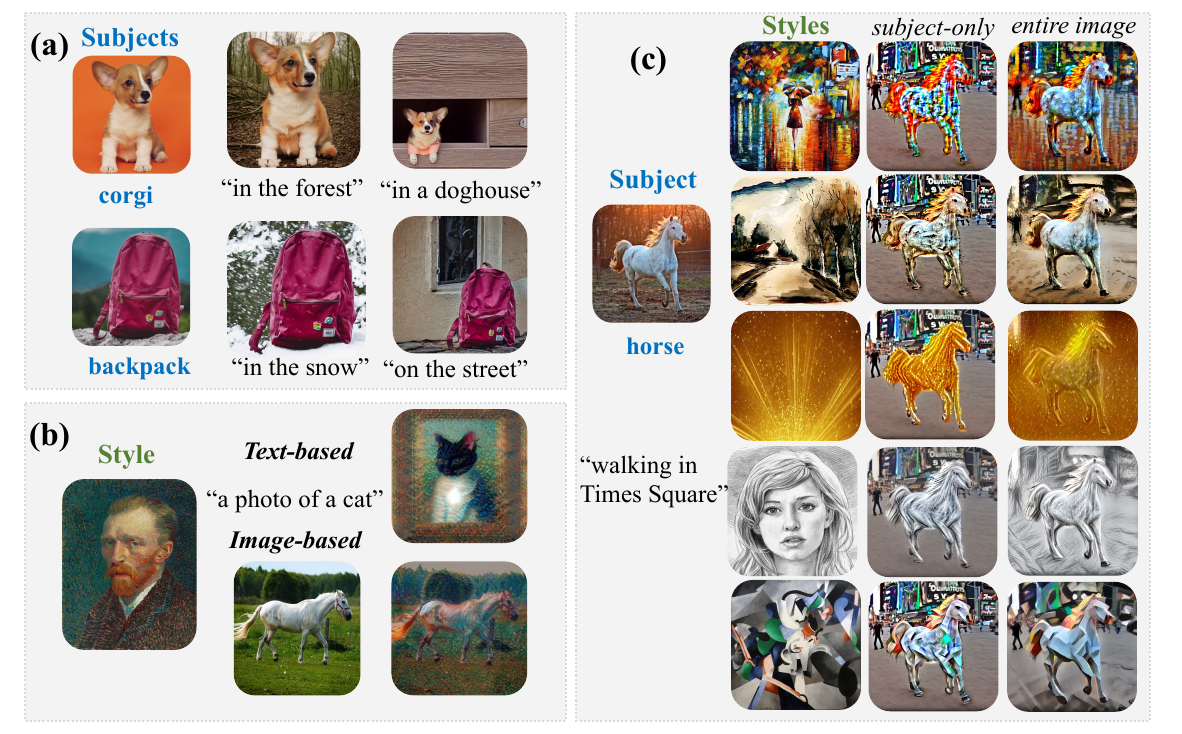}
    \vspace{-0.5em}
    \caption{\small Given a subject image and a style image, our training-free method FreeTuner can  support various personalized image generation: \textbf{(a)} subject-driven, \textbf{(b)} style-driven, and \textbf{(c)} compositional personalization.}
    \label{fig:main}
\end{figure*}

\begin{abstract}
With the advance of diffusion models, various personalized image generation methods have been proposed. However, almost all existing work only focuses on either subject-driven or style-driven personalization. Meanwhile, state-of-the-art methods face several challenges in realizing \textbf{compositional personalization}, \ie, composing different subject and style concepts, such as concept disentanglement, unified reconstruction paradigm, and insufficient training data. To address these issues, we introduce \textbf{FreeTuner}, a flexible and training-free method for compositional personalization that can generate \emph{any user-provided subject in any user-provided style} (see \Fig~\ref{fig:main}). Our approach employs a disentanglement strategy that separates the generation process into two stages to effectively mitigate concept entanglement. FreeTuner leverages the intermediate features within the diffusion model for subject concept representation and introduces style guidance to align the synthesized images with the style concept, ensuring the preservation of both the subject's structure and the style's aesthetic features. Extensive experiments have demonstrated the generation ability of FreeTuner across various personalization settings.
\end{abstract}
 
\section{Introduction}
Recently, diffusion models ~\cite{ Dalle2,Dalle,ldm,  Imagen} have demonstrated impressive superiority in the realm
of image generation. Owning to their unprecedentedly creative capabilities, an emerging trend,
personalized image generation~\cite{Dreambooth,textInversion,ip-adapter} has attracted much interest due to its broad applications in daily life such as art creation, advertising, and entertainment. Within these innovative applications, users can create images that adhere to user-specific visual concepts\footnote{In this paper, we regard different \emph{objects} or \emph{styles} as different ``concepts''.}.
As shown in Figure~\ref{fig:main}, current personalized generation work can be roughly divided into two directions: 1) \textbf{Subject-driven}~\cite{Dreambooth,SSREncoder, BLIPDiffusion}: They aim to synthesize photorealistic images of the user-provided subjects in a novel context based on text prompts. \eg, we can generate the user-provided \texttt{corgi} in various new scenarios (\cf, Figure~\ref{fig:main}(a)). 2) \textbf{Style-driven}~\cite{instantstyle,deadiff,ip-adapter}: They aim at generating the image to follow the reference style while preserving its content. As shown in Figure~\ref{fig:main}(b), this kind of personalization includes text-based stylization and image-based stylization. 

Subsequently, various types of personalization methods have been proposed: 1) \emph{Test-time fine-tuning}: They generally utilize an optimized placeholder text embedding~\cite{textInversion} or fine-tune the pre-trained model with different regularizations~\cite{Dreambooth} to learn the user-provided concept. 2) \emph{Adapter-based}~\cite{BLIPDiffusion,deadiff}: They typically train an additional encoder and then map the concept image into the image embedding to guide the generation process.
However, 
\textcolor{black}{almost all existing work only focuses on either subject-driven or style-driven personalization, without considering the \textbf{compositional personalization} (\ie, a specific subject portrayed in a specific style)}. For example in Figure~\ref{fig:main}(c), artists may want to synthesize an image with the \texttt{horse} in a new scenario (\eg, \texttt{walking in Times Square}) \textcolor{black}{and wish the horse or even the entire image is rendered in a unique style to spark their creativity.}

Despite the increasing demand,  previous methods~\cite{Dreambooth,textInversion,ip-adapter,BLIPDiffusion,deadiff,style_inj,styledrop} face several challenges in effectively composing different subject and style concepts: 1) \textbf{Concept Disentanglement.} The relationship between style and subject concepts is intricately entangled~\cite{deadiff,BreakforMakeML,instantstyle}. Previous methods lack effective strategies to decouple them, which confuses the diffusion model and makes it difficult to distinguish between subject and style concepts during the generation process. 2) \textbf{Unified Reconstruction Paradigm.} Both tuning-based and adapter-based methods require a similar objective function to reconstruct the concept within the same parameter space. This unified training paradigm makes the entanglement problem even worse. 3) \textbf{Insufficient Training Data.} Tuning-based methods such as DreamBooth~\cite{Dreambooth} require a collection of images for each concept (\eg, 3-5 images), while adapter-based methods need a larger scale of image collection. Additionally, to combine subject and style concepts, adapter-based methods~\cite{ip-adapter,BLIPDiffusion} need to collect large amounts of subject-style image pairs to train the encoder. However, due to the indeterminate definition of style~\cite{instantstyle}, collecting images of the same style is difficult, let alone images combining the same subject with the same style.





\begin{wrapfigure}[13]{tr}{0.4\textwidth}
    \vspace{-2em}
    \centering
    \includegraphics[width=1\linewidth]{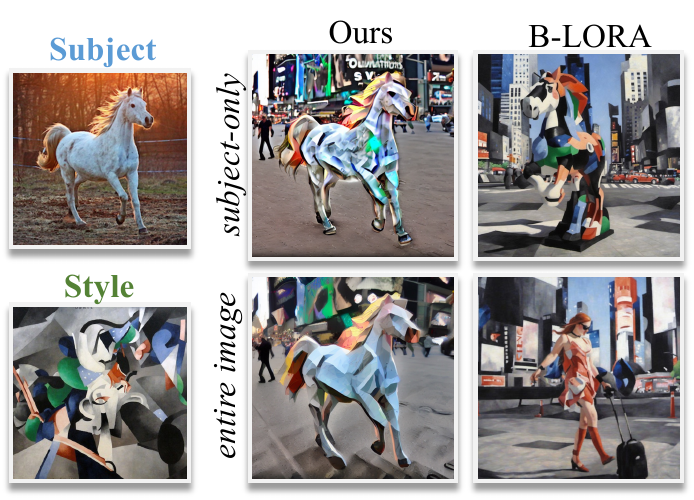}
    \captionsetup{font=small}
    \captionsetup{margin=0.3em}
    \vspace{-1.5em}
    \caption{\small Given \texttt{``A photo of a horse walking in Times Square"}, B-LoRA not only distorts the horse's structure but also fails to render the entire scene.}
    \label{fig:compare_with_blora}
\end{wrapfigure}


In pursuit of compositional personalization, few recent methods~\cite{DBlora, ZipLoRAAS, B-lora, BreakforMakeML} introduce multiple LoRAs~\cite{lora} to decouple the image, such as B-LoRA~\cite{B-lora}. 
It requires only one image of a concept and it employs LoRA on different layers of SDXL~\cite{sdxl} to represent the image's content and style separately, partially mitigating disentanglement issues. 
However, the intricate process of layer-wise LoRA tuning requires significant computational resources with a substantial amount of time. Furthermore, it disrupts the structural information of the subject concept and it can only associate the style concept with a single subject concept, rather than personalize the entire generated image with the reference style (see \Fig~\ref{fig:compare_with_blora}), which greatly limits its application scopes.

\begin{wrapfigure}[12]{t}{0.6\textwidth}
    \centering
    \includegraphics[width=1\linewidth]{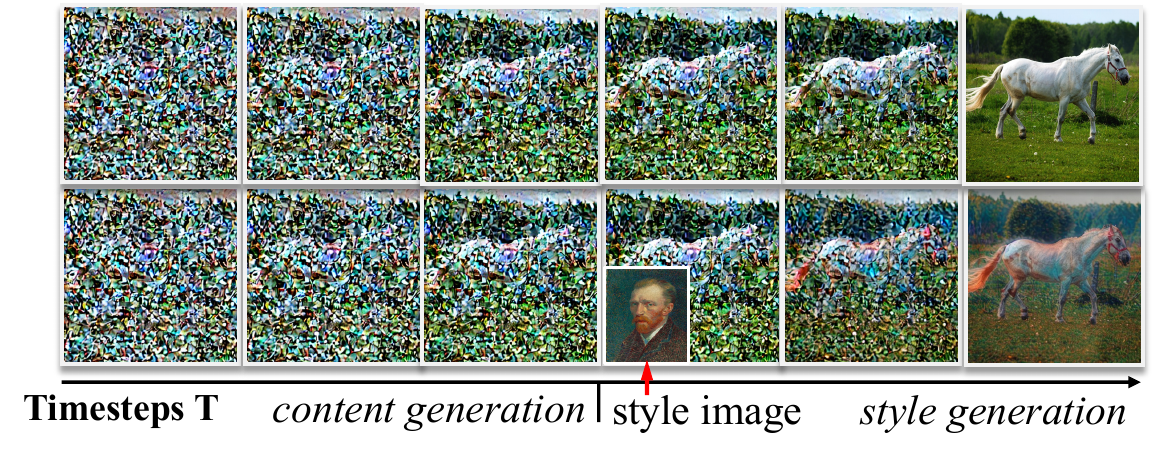}
    \captionsetup{font=small}
    \captionsetup{margin=0.3em}
    \vspace{-2em}
    \caption{\small Visualization of the image estimations corresponding to different timesteps within the denoising process (\texttt{row 1}) and our two-stage disentanglement strategy (\texttt{row 2}).}
    \label{fig:mid_state}
\end{wrapfigure}

\textcolor{black}{To address the aforementioned challenges in compositional personalization while reducing the computational cost}, we present \textbf{FreeTuner}, a versatile training-free method based on diffusion models that only requires one image for each concept. 
\textcolor{black}{FreeTuner is built on the premise that the diffusion model generates an image in a coarse-to-fine manner~\cite{localizing,P+}. For example in \Fig~\ref{fig:mid_state}, the rough content of the image is generated first, and then fine-grained details follow. 
Inspired by this,} \\
FreeTuner adopts a simple but effective disentanglement strategy that divides the generation process into two stages along denoising steps: 1) \emph{Content generation stage}: It focuses on the generation of subject concepts. 2) \emph{Style generation stage}: It aims to synthesize the features of style concepts such as tones and textures. This division strategy explicitly separates the subject concept generation from the style concept generation, thereby mitigating the entanglement problem during the generation process. Specifically, for content generation,
we utilize the intermediate features (\eg, attention maps) within the diffusion model to generate rough content of the subject concept. For style generation, we introduce style guidance to penalize discrepancies between the predicted synthesized and style concept images, effectively steering the generation process towards a similar style expressed in the style concept. By injecting intermediate features into the content generation stage and employing style guidance in the style generation stage, FreeTuner ensures that both the structural integrity of the subject and the aesthetic characteristics of the style are preserved, resulting in a harmonious blend of different concepts.

\textcolor{black}{FreeTuner offers a significant advantage over training-based approaches by eliminating the need for training additional encoders or fine-tuning the pre-trained diffusion models}. To the best of our knowledge, it is the first training-free method capable of subject-style compositional personalization, thanks to its effective decoupling strategy. Extensive experiments have demonstrated that FreeTuner achieves state-of-the-art performance across various concept personalization settings.


In summary, our contributions are as follows:
 
\vspace{-1em}

\begin{itemize}[leftmargin=*]
\itemsep-0.2em

\item we propose FreeTuner, a training-free method for compositional personalization, requiring only one image for each concept.


\item We propose a decoupling strategy, which effectively solves the subject-style concept entanglement problem by explicitly separating the subject concept generation from the style concept generation.

\item Our method presents the first universal training-free solution that supports various personalized image 
generation (multi-concept, subject-driven, style-driven) and controllable diffusion models. 

\end{itemize}

    

\section{Related Work}

\noindent\textbf{Subject-driven Personalization.} Current subject-driven methods can be categorized into two types: \emph{1) Test-time fine-tuning}~\cite{textInversion, P+, Dreambooth, Customization, break-a-scene,disenbooth}: They typically use an optimized placeholder text embedding or fine-tune the pre-trained model to learn the user-provided subject. For example, Textual Inversion~\cite{textInversion} optimizes an additional text embedding for representing a new subject, while $P+$~\cite{P+} optimizes multiple embeddings to enhance its expressive capability and precision. DreamBooth~\cite{Dreambooth} adjusts the weight of the diffusion U-net to associate new subjects with unique identifiers. \emph{2) Tuning-free methods}~\cite{Elite, BLIPDiffusion, taming_Encoder, Encoder-based-domain, SSREncoder}: They generally train an additional encoder on large-scale datasets to map the subject image into image embedding for subject-driven generation. For instance, ELITE~\cite{Elite} trains an encoder, which supports global and local mapping for subject-driven generation. BLIP-Diffusion~\cite{BLIPDiffusion} pre-trains a multimodal encoder to enable efficient fine-tuning or zero-shot subject-driven generation. SSR-Encoder~\cite{SSREncoder} trains a novel encoder to support selective subject-driven generation. Although these methods can generate customized images of subjects, their time-consuming training or tuning process significantly hinders their usage in practical applications.


\noindent\textbf{Style-driven Personalization.}
For style-driven personalization, previous methods~\cite{Dreambooth, textInversion, P+, lora} require the collection of a set of images sharing the same style, and then learn the style concepts by reconstructing them. While  DEADiff~\cite{deadiff} utilizes the generated data synthesized by a state-of-the-art text-to-image model Midjourney
with text style descriptions to train an additional image encoder. Considering the inherently intricate nature of the visual style, building such datasets is labor-intensive and restricted to the number of styles, leading to a bottleneck for applications in practice. Recently, inversion-based methods StyleAlign~\cite{StyleAligned} and StyleInj~\cite{style_inj} have designed fusion operations on intermediate features between user-provided style image reconstruction streams and other streams. Nevertheless, these methods involve inverting the style image to obtain intermediate features which may result in loss of fine-grained style components such as color tone and texture. In this paper, we draw inspiration from traditional Neural Style Transfer methods~\cite{style_transfer_0, style_transfer_1, style_transfer_2, style_transfer_3, adain, WikiArt, style_transfer_4, style_transfer_5}. Specifically, we introduce pre-trained networks~\cite{vgg19} along with a guidance function to direct the denoising process towards a given style. This approach allows for seamless incorporation of style guidance into the denoising step without requiring additional training.

\noindent\textbf{Compositional Personalization.} It aims to generate an image that preserves both the structure of the subject and the aesthetics of the style while aligning well with the text prompt. Recent methods~\cite{DBlora, ZipLoRAAS, B-lora, BreakforMakeML} have attempted to achieve this innovative idea. They typically utilize multiple LoRAs~\cite{lora} to capture subject and style separately, and then employ different strategies for combining them.  For instance, a common approach~\cite{DBlora} is to combine LORAs by assigning different weights. ZipLoRA~\cite{ZipLoRAAS} has devised a complex fusion strategy that merges two individual LoRAs trained for style and subject into a new  ``zipped'' LoRA. B-LoRA~\cite{B-lora} proposes a layer-wise LoRA tuning pipeline that utilizes LoRA on different layers of SDXL~\cite{sdxl} to represent an image’s content and style respectively. However, all these methods face challenges in efficiently disentangling content and style due to the unified training paradigm, which is also time-consuming.
\section{Method}

\subsection{Preliminaries}
\label{sec:pre}
\textbf{Latent Diffusion Model (LDM).}
LDM~\cite{ldm} consists of an encoder $\mathcal{E}$ and decoder $\mathcal{D}$ trained with a reconstruction objective. Given an image $x$, encoder $\mathcal{E}$ projects $x$ into a latent code  $z$ and decoder $\mathcal{D}$ reconstructs the image from the latent code, \ie, $ \widetilde{x}= \mathcal{D}(z) = \mathcal{D}(\mathcal{E}(x)).$  With the pre-trained encoder, they project each image into a latent space $z$, and then train a diffusion model on $z$ by predicting noise $\hat{\epsilon} = \epsilon_\theta(z_t,t,y)$ conditioned on any timestep $ t\in \{0,..., T\}$ and an additional signal like text prompt $y$.
The diffusion model is trained by minimizing the  denoising score matching objective~\cite{ddpm}:
\begin{equation}
\label{eq:ldm_training_objective}
\mathcal{L}=\mathbb{E}_{z\sim\mathcal{E}(x),y,\epsilon\sim\mathcal{N}(0,1),t}\left[\|\epsilon-\epsilon_\theta(z_t;t,y)\|_2^2\right].
\end{equation}
Here, $z_t=\sqrt{\bar{\alpha}_t}z+\sqrt{1-\bar{\alpha}_t}\epsilon$, $\bar{\alpha}$ is a predefined noise adding weight and $\epsilon_\theta$ is a denoising network.
By removing predicted noise from  $z_t$, we can obtain a cleaner latent code $z_{t-1}$, we denote $z_{t-1} = DM(z_t,t,y)$ as one denoising step in this paper.

\noindent\textbf{Attention Mechanisms in Denoising Network $\epsilon_\theta$.} Typically, $\epsilon_\theta$ is a U-Net architecture including both self-attention and cross-attention mechanisms. For self-attention maps, they are computed as $SA=Softmax(\frac{Q_sK_s^T}{\sqrt{d}})$, where $Q_s$ and $K_s$ represent different projections of visual features. For cross-attention maps, they can be calculated by $CA=Softmax(\frac{Q_cK_c^T}{\sqrt{d}})$, where $Q_c$ denotes the projection of textual embedding and $K_c$ denotes the projection of visual feature.

\noindent\textbf{Guidance Diffusion.}
Classifier guidance~\cite{classifier_g} utilizes a noise-dependent external classifier to modify the sampling process. Actually, any measurable object properties can serve as an energy function $g(z_t;t,y)$ to guide the sampling process~\cite{DiffusionSelf}, including layout control through attention maps~\cite{boxdiff} or appearance guidance~\cite{freecontrol}, and it even can be incorporated with classifier-free guidance~\cite{classifier_f}:
\begin{equation}
\label{eq:class_free}
    \hat{\epsilon}_t=(1+s)\epsilon_\theta(z_t;t,y)-s\epsilon_\theta(z_t;t,\emptyset)+v\sigma_t\nabla_{z_t}g(z_t;t,y),
\end{equation}
where $s$ is a parameter that controls the strength of the classifier-free guidance, and  $v$ is an additional guidance weight for the energy function $g(\cdot)$.

\subsection{FreeTuner}
\label{sec:overview}

\noindent\textbf{Task Formulation\footnote{For presentation simplicity, we only show single-subject and single-style composition personalization here. However, our FreeTuner can be easily extended to multiple subjects or styles scenarios (\cf, \Fig~\ref{fig:other_uses_case}(a)).}.}
Given a subject image $I_{sub}$, a style image $I_{sty}$, and a text prompt $P_{comp}$, we aim to synthesize an image $I_{comp}$ that satisfies the description of $P_{comp}$. We can render either an entire $I_{comp}$ or just the subject within $I_{comp}$ as the style of $I_{sty}$. Moreover, we can also flexibly control the location $l$ of the subject in $I_{comp}$. 

\begin{figure*}[t]
    \centering
    \includegraphics[width=1.0\textwidth]{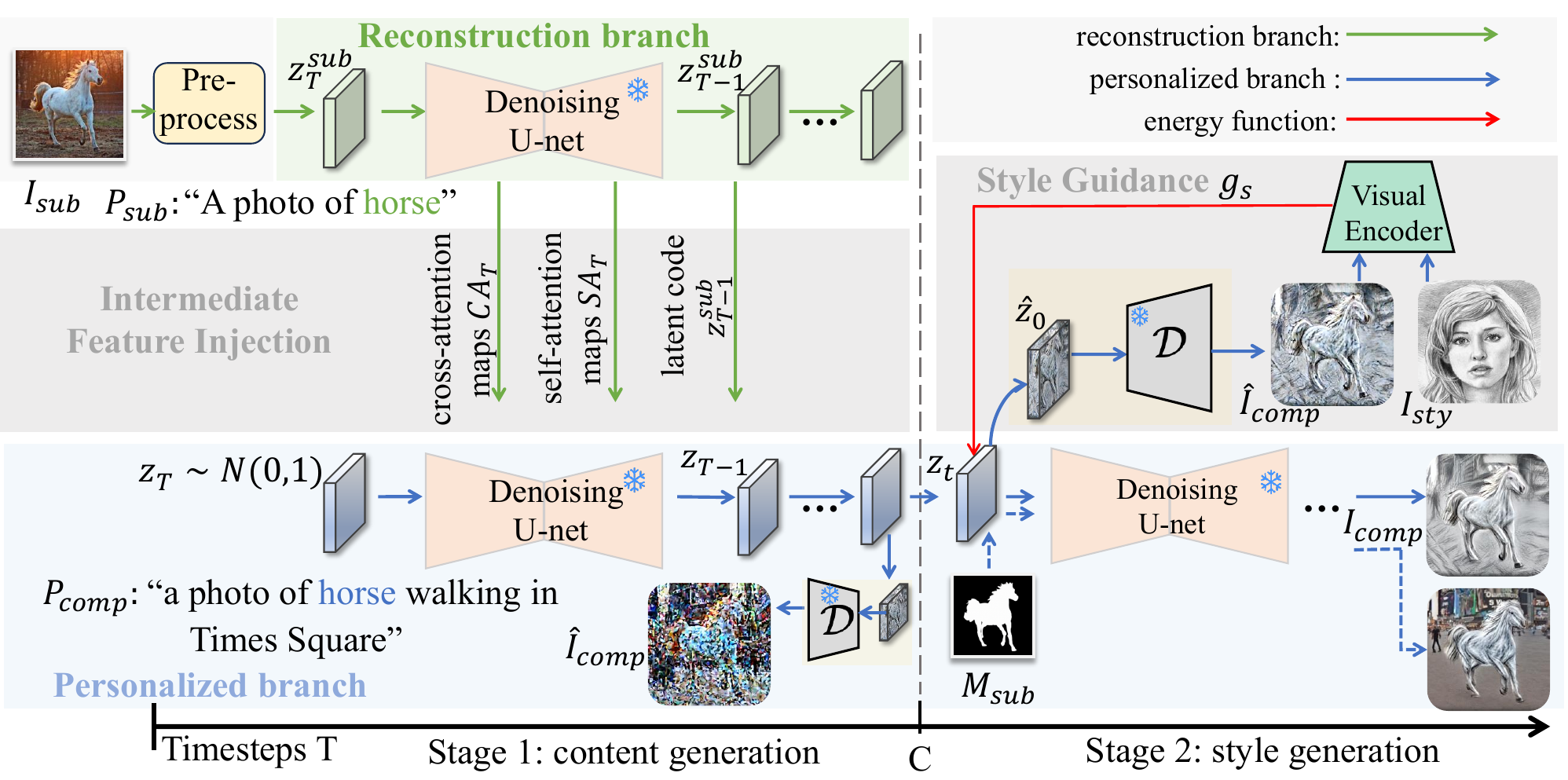}
    \vspace{-1em}
    \caption{\small\textbf{Overview of the proposed FreeTuner.} (a) In the preprocessing step, we get a binary mask $M_{sub}$ including only the subject through off-the-shelf models and inverse $I_{sub} * M_{sub}$ with a simple prompt $P_{sub}$ to acquire latent code $z_T^{sub}$. (b) Our generation process is divided into two stages. In the first stage, we focus on content generation which injects the intermediate features obtained from the reconstruction branch into the personalized branch.
    Upon entering the style generation stage, 
    an additional visual encoder (\eg, VGG-19~\cite{vgg19}) and guidance function will steer the generated image toward a similar style expressed in $I_{sty}$.}
    \label{fig:pipe}
\end{figure*}

\noindent\textbf{Pipeline Overview.}
As illustrated in \Fig~\ref{fig:pipe}, our proposed FreeTuner consists of two stages: \textbf{1)} Content Generation Stage: The \emph{intermediate features}\footnote{The ``intermediate features'' consist of cross-attention maps, self-attention maps, and latent codes. \label{footnote:feature}} from the reconstruction branch are utilized to generate the coarse-grained content of the subject. \textbf{2)} Style Generation Stage: It focuses on fine-grained detail generation such as tones and texture. \textcolor{black}{An additional visual encoder (\eg, VGG-19~\cite{vgg19}) and energy function will steer the generated image toward a similar style expressed in $I_{sty}$.}

\noindent\textbf{Subject Preprocessing.} Following previous subject-driven personalization generation work~\cite{mudi}, for each subject image $I_{sub}$, we have a subject preprocessing step. Specifically, for $I_{sub}$ and its corresponding class name, we can get a binary mask of the subject  $M_{sub}$ and a handcrafted prompt $P_{sub}$ containing its class name, \eg, \texttt{``a photo of horse''}. Subsequently, we inverse $M_{sub} * I_{sub}$ with prompt $P_{sub}$ to get the initial latent code $z^{sub}_T$ of the subject. (\textcolor{black}{More details are in the Appendix}.)

\subsubsection{Stage 1: Content Generation}
\label{sec:content_methods}
\textcolor{black}{Given a random Gaussian noise $z_T$ and the prompt $P_{comp}$, this stage aims to generate an intermediate latent code, which has the coarse-grained content information of the given subject.}
The key ideas are leveraging the intermediate features\footref{footnote:feature} from the reconstruction branch.

\noindent\textbf{Subject-related Feature Injection.} Upon obtaining the latent code $z_T^{sub}$, we first reconstruct the subject through the denoising step: 
$z_{t-1}^{sub} = DM(z_t^{sub},t,P_{sub})$.
In each denoising step, we can get the latent codes of the subject $z_{t-1}^{sub}$, along with the self-attention maps $SA_t$ and cross-attention maps $CA_t$ consisting of $N$ attention layers. Every word in $P_{sub}$ corresponds to an attention map $CA_t^i$. These intermediate features\footref{footnote:feature} have been widely recognized to contain valuable information about the content and layout of the subject image~\cite{p2p,masactrl,plug_play}. Thus, we inject them into the content generation stage to preserve the visual appearance of the subject with three following feature swap operations.

\noindent\emph{\underline{1) Cross-attention Map Swap.}}
Inspired by the image editing method~\cite{p2p}, we selectively swap the subject-related CA maps, while keeping the others unchanged, to ensure semantic coherence between the generated subject and the user-provided subject.
\begin{equation}
\widetilde{CA}_{t}^{i*} = 
\begin{cases}
CA_{t}^i, & \text{if } t \leq \tau \text{ and } w_i \text{ in } P_{sub}\\
\widetilde{CA}_{t}^i, & \text{otherwise},
\end{cases}
\label{swap_CA}
\end{equation}
where $\widetilde{CA}_t$ denotes the CA maps in the personalized branch denoising step, $\tau$ is a timestamp hyperparameter that determines which step the swap is applied and $w_i$ is a word in $P_{comp}$. To achieve a better balance between image personalization and reconstruction, we only swap the CA maps in the first few timestamps but rather utilize the SA maps as discussed below.

\noindent\emph{\underline{2) Self-attention Map Swap.}}
The SA mechanisms in the diffusion model have been demonstrated to have a potent correlation with the spatial layout~\cite{masactrl,plug_play}. To achieve fine-grained control of the overall generated content, while minimizing the impact on the personalization of the subject. We only swap the subjected-related region in SA maps while keeping the others unchanged:
\begin{equation}
\label{swap_SA}
\widetilde{SA}_t^{*} = SA_t * M_{sub} + \widetilde{SA}_t * (1 - M_{sub}).
\end{equation}
Here, $\widetilde{SA}_t$ is the SA maps in the personalized denoising step.

\noindent\emph{\underline{Attention Map Summary.}} After swapping CA and SA maps, the personalized denoising step becomes: 
\begin{equation}
\label{cust_denoise_step}
z_{t-1} = DM^{*}(z_t,t,P_{comp}; \widetilde{SA}_t^{*}, \widetilde{CA}_t^{*}).
\end{equation}
where we use $DM^{*}$ to denote the modified denoising step with the changed attention maps\footnote{Unlike image editing methods~\cite{p2p}, they usually start with an inversed latent code and swap intermediate features during the denoising process to achieve real image editing. FreeTuner starts with random Gaussian noise and swaps only the attention maps relevant to the subject, ensuring high-quality subject personalization.}.

\noindent\emph{\underline{3) Latent Codes Swap.}}
Inspired by the latent blending strategy~\cite{blend, Blendedldm} for achieving user-specified region editing, we argue that the latent codes include valuable information about the content of the generated image. To further keep the fine-grained visual appearance of the subject while aligning with the prompt $P_{comp}$, we perform subject-related latent codes swap:
\begin{equation}
\label{swap_latent}
z_{t-1} = z_{t-1} * M_{sub} + z_{t-1}^{sub} * (1 - M_{sub}).
\end{equation}

\begin{wrapfigure}[12]{tr}{0.6\textwidth}
    \vspace{-2em}
    \centering
    \includegraphics[width=0.95\linewidth]{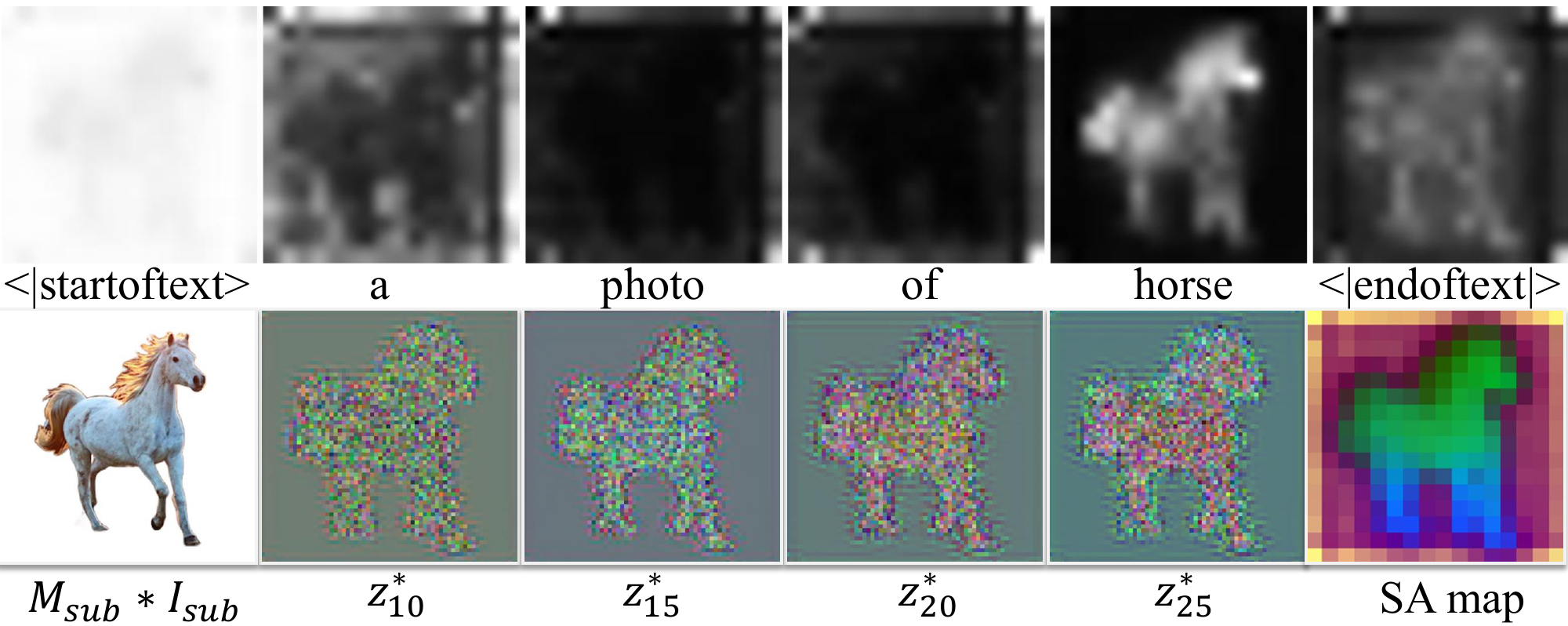}
    \captionsetup{font=small}
    \captionsetup{margin=1em}
    \vspace{-1pt}
    \caption{\small\textbf{Visualization of subject-related features}: The top row displays the average CA maps for each word in the $P_{sub}$. In the bottom row, we perform PCA on latent codes $z$ across all diffusion steps and SA maps.}
    \label{fig:vis_fea}
\end{wrapfigure}

As shown in \Fig~\ref{fig:vis_fea}, we visualize the leading principal components of the latent codes along the diffusion steps, finding that the latent codes are visually equivalent to the generated image. Note to prevent a simple duplication of the subjects, we perform Eq.~\eqref{swap_latent} only in a few timestamps.

\noindent\textbf{Spatial-constrained Strategy.}
While the above-mentioned feature injection can achieve a photorealistic generation of the subject, pixel-level artifacts still occur. The reasons are that our personalized branch starts from a random noise and it is conditioned on $P_{comp}$ rather than $P_{sub}$. To address this issue, we propose a spatial-constrained strategy to better align the visual appearance of the subject in the latent space. By updating the Eq.~\eqref{eq:class_free} into:
\begin{equation}
\hat{\epsilon}_t=(1+s)\epsilon_\theta(z_t;t,P_{comp})-s\epsilon_\theta(z_t;t,\emptyset)+\lambda_l\mathcal{L}(M_{l},CA_{t}^{sub}),
\label{spatial_constrain}
\end{equation}
where the energy function $\mathcal{L}$ guides the model to focus specifically on the subject, $\lambda_l$ is the guidance strength, $M_l$ is a binary mask transformed from the top-left and bottom-right coordinates of user-provided location $l$, and $CA_t^{sub}$ is the cross-attention map of the subject word.
Our spatial-constrained strategy is building on the methodology presented by BoxDiff~\cite{boxdiff}, which adopts Inner-Box, Outer-Box, and Corner Constraints to achieve a training-free layout-to-image generation. Thus our energy function can be expressed as : \textcolor{black}{$\mathcal{L} = \mathcal{L}_{IB} + \mathcal{L}_{OB} + \mathcal{L}_{CC}$}\footref{ft:appendix}.
 
\subsubsection{Stage 2: Style Generation}
\label{sec:style_methods}

After the content generation stage, we can get the intermediate latent code $z_c$ which includes the coarse-grained visual information of the subject. In this stage, our target is to update the latent code towards the style image $I_{sty}$. However, finding appropriate updating directions is challenging due to the absence of measurable style properties in latent space. To address this, we provide specific guidance on the estimation of the final result $I_{comp}$ in the pixel space.The estimation of $I_{comp}$  can be derived from the current noised latent $z_{t}$ and the model’s noise prediction by decoder $\mathcal{D}$ via:
\begin{equation}
\label{predic_x0}
\hat{z}_0=\frac{z_{t}-\sqrt{1-\bar{\alpha}_{t}}\epsilon_\theta(z_{t}; t, P_{comp})}{\sqrt{\bar{\alpha}_{t}}}, \quad \hat{I}_{comp}=\mathcal{D}(\hat{z}_0).
\end{equation} 
\noindent\textbf{Style Guidance.}
Inspired by previous style transfer methods~\cite{style_transfer_0,style_transfer_1,adain}, we utilize a pre-trained visual encoder (\eg, VGG-19~\cite{vgg19}) as an external supervisor to penalize the difference between the predicted image $\hat{I}_{comp}$ and style image $I_{sty}$. We express the energy function $g_s$ for style guidance:
\begin{equation} \label{style_loss}
\small
g_{s}(\hat{I}_{comp};I_{sty})=\textstyle{\sum}_{i=1}^L \left[ \lVert\mu(f_i(B(\hat{I}_{comp})))-\mu(f_i(I_{sty}))\rVert_2 + \lVert\sigma(f_i(B(\hat{I}_{comp})))-\sigma(f_i(I_{sty}))\rVert_2 \right],
\end{equation}
where $f_i$ symbolizes the i-th layer in the VGG-19 model, $B$ is the bi-linear interpolation operation, $\mu$ and $\sigma$ represent the mean and standard deviation of the features respectively.

\noindent\textbf{Content Preservation Guidance.} Although style guidance can achieve a high-quality style-driven personalization, we find it may destroy the content of the generated image. To get a balance between style-driven personalization and content preservation, we adopt the content preservation guidance: 
\begin{equation}
\label{content_loss}
g_c=\|F(\hat{I}_{comp})-\text{AdaIN}(F(M_{sub}*I_{sub}),F(I_{sty}))\|_2,
\end{equation}
where  $F$ is a set of $f_i$, $\text{AdaIN}$ is the Adaptive Instance Normalization~\cite{adain}.

\noindent\textbf{Guiding the Style Generation Process.}
We can update Eq.~\eqref{eq:class_free} by incorporating $g_s$ and $g_c$:
\begin{equation}
\label{style_generation_guidacne}
\hat{\epsilon}_t=(1+s) \epsilon_\theta(\mathbf{z}_t;t,P_{comp})-s \epsilon_\theta(\mathbf{z}_t;t,\emptyset)+\lambda_s g_s+\lambda_c
 g_c,
\end{equation}
where $\lambda_s$ and $\lambda_c$ are the guidance strengths. As shown in \Fig~\ref{fig:mid_state}, our style generation stage can achieve a high-quality style personalization while preserving the fine-grained detail of the content perfectly. It is worth noting that all pretrained models are frozen and our method can be easily incorporated with other diffusion models.

\begin{figure*}[t]
\centering
\includegraphics[width=0.88\textwidth]{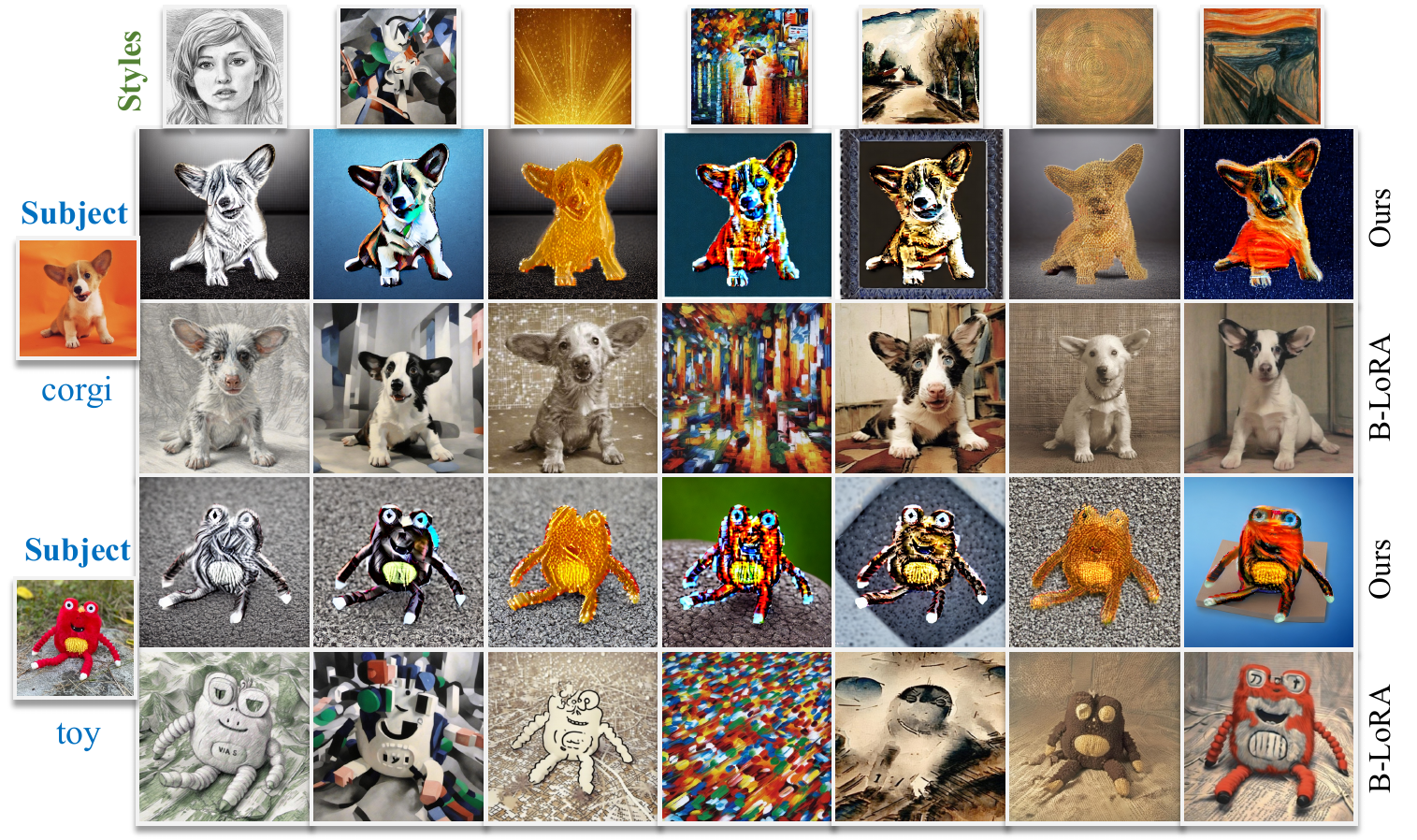}
\vspace{-1em}
\caption{\small\textbf{Qualitative Comparison on single subject-style composition personalization.} We compare FreeTuner and B-LoRA with template \texttt{``A photo of a [class name]"} for generation.
}
\vspace{-0.5em}
\label{fig:any_subject_any_style}
\end{figure*}

\begin{figure*}[t]
    \centering
    \includegraphics[width=0.98\textwidth]{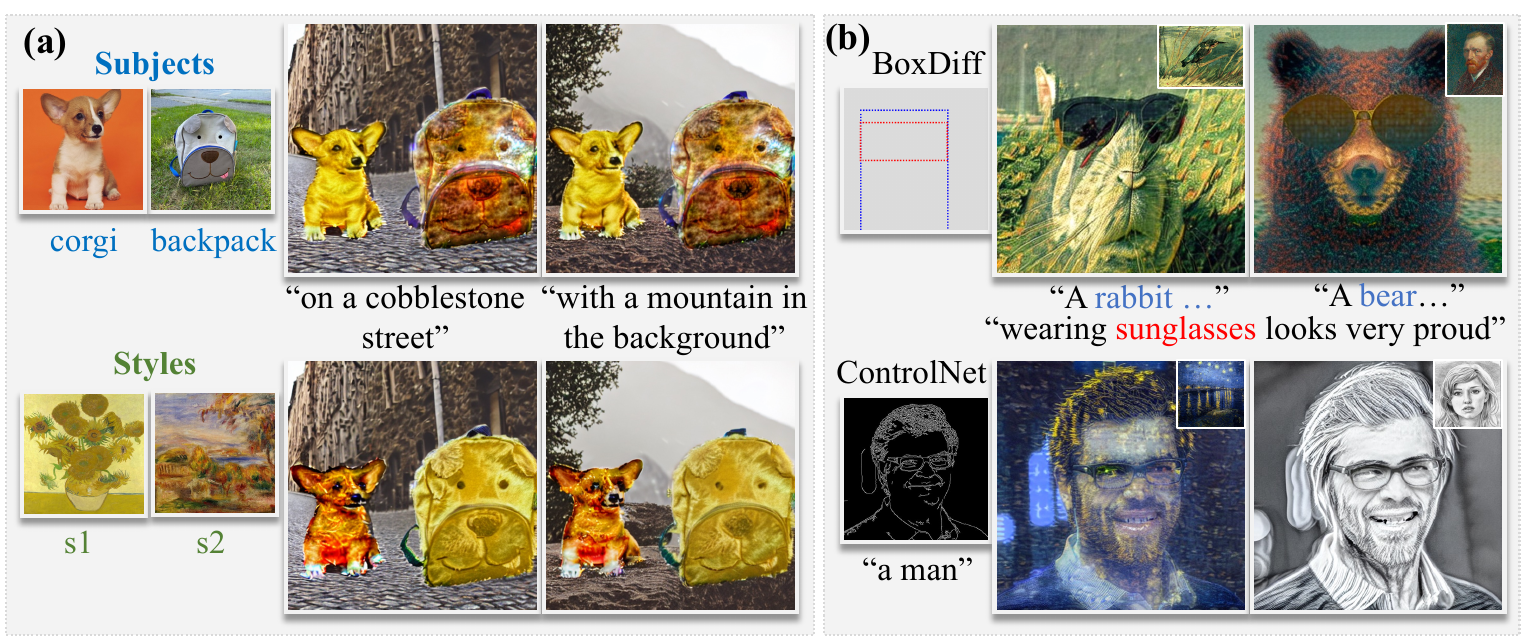}
    \vspace{-0.5em}
    \caption{\small Results of \textbf{(a): Multiple Subject-Style Personalization.} FreeTuner can personalize multiple subjects and styles with different combinations.
    \small\textbf{(b): Combined with other diffusion-based methods}. On top, our style guidance is combined with the training-free method BoxDiff~\cite{boxdiff} to transfer the style. On the bottom, the content is synthesized by our style guidance and ControlNet~\cite{ControNet} conditioned on the sketch.
    }
    \vspace{-1.5em}
    \label{fig:other_uses_case}
\end{figure*}

\begin{figure*}[t]
    \centering
    \includegraphics[width=0.98\textwidth]{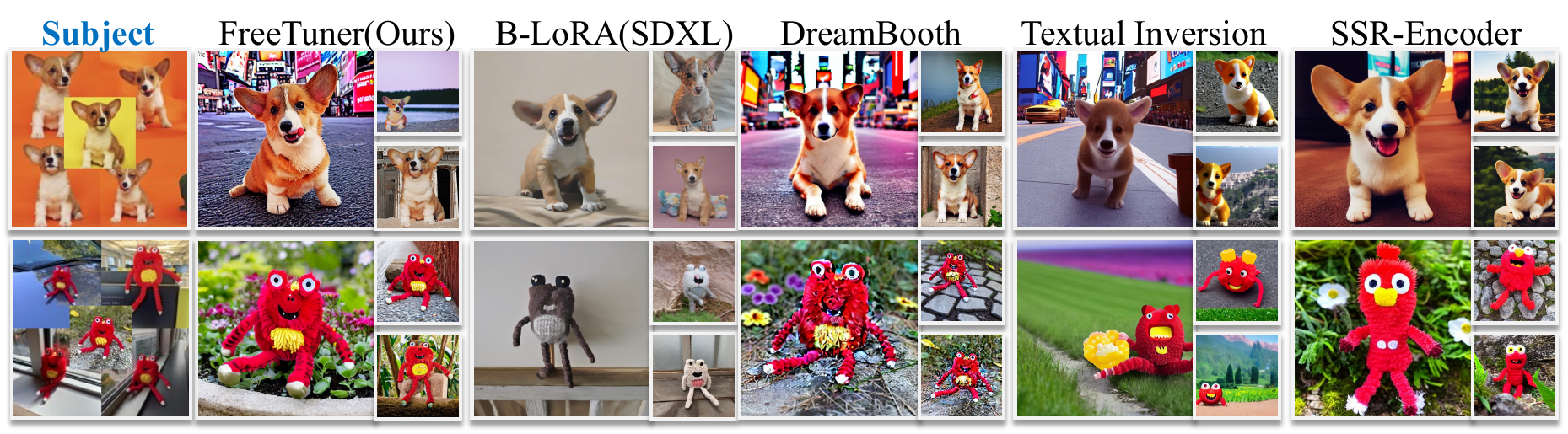}
    \vspace{-0.5em}
    \caption{\small\textbf{Qualitative comparison on subject-driven personalization}. For B-LoRA, a simple prompt \texttt{``a [class name]"} is used for the generation, while others use the same detailed prompts (\cf. Appendix).}
    \vspace{-0.5em}
    \label{fig:single_subject}
\end{figure*}

\begin{figure*}[t]
\centering
\includegraphics[width=0.95\textwidth]{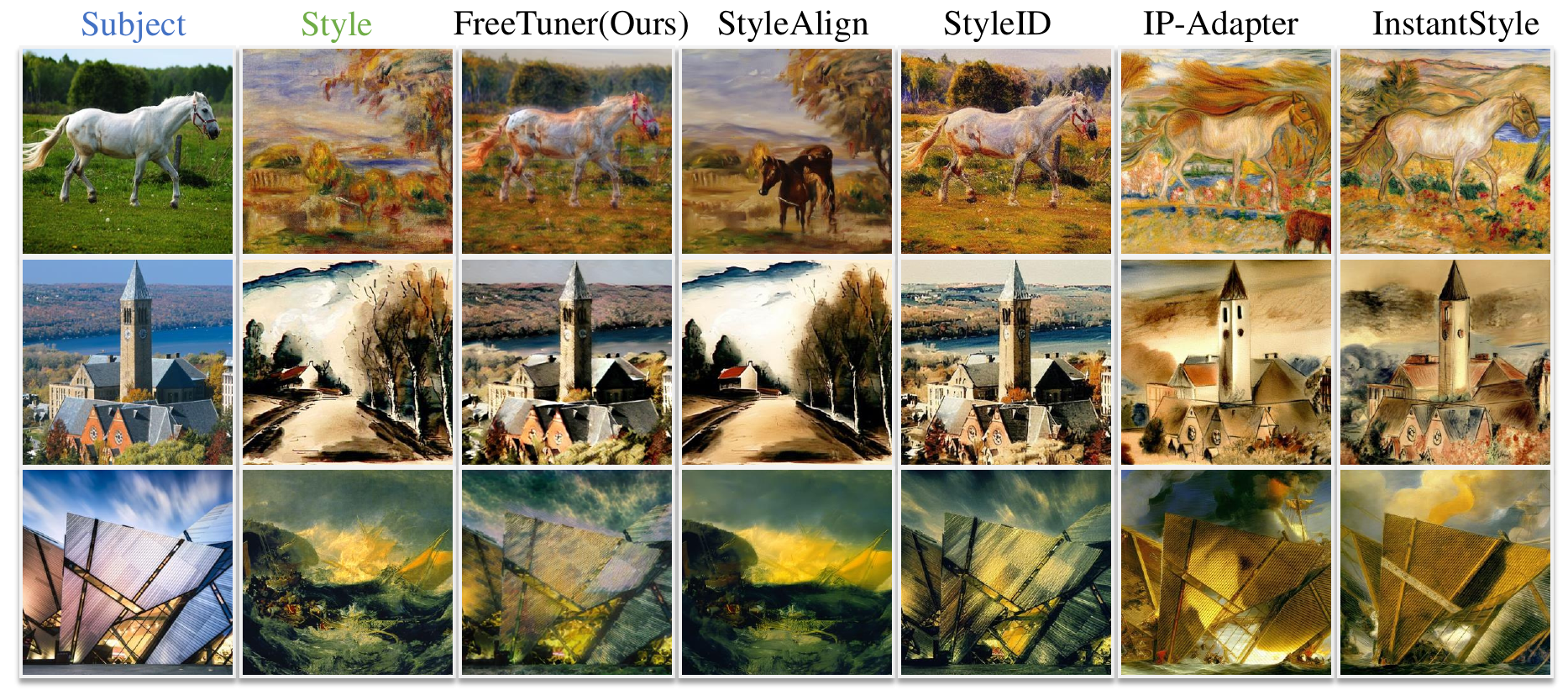}
\vspace{-1ex}
\caption{\small\textbf{Qualitative comparison on style-driven personalization}.
}
\vspace{-1.5em}
\label{fig:style_compare}
\end{figure*}

\begin{figure*}[t]
    \centering
    \includegraphics[width=1.0\textwidth]{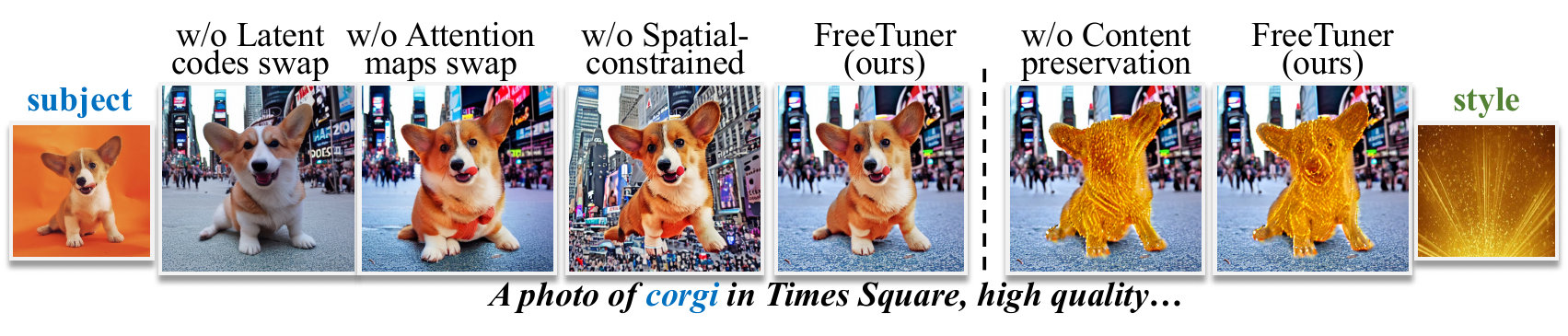}
    \vspace{-2em}
    \caption{\small Ablation study on proposed components in content generation (Left) and style generation (Right).  
    }
    \label{fig:ablation_study}
\end{figure*}

\section{Experiments}
\noindent\textbf{Dataset.}
We evaluated FreeTuner with a diverse set of subject images from ~\cite{Dreambooth}, which contains 30 subjects each depicted by 4-5 images. We employed style images from StyleDrop~\cite{styledrop} and WikiArt~\cite{WikiArt}.
\textbf{Implementation Details.}
We implemented our method on Stable Diffusion V1.5. We used null-text inversion ~\cite{null_textInversion} based on DDIM inversion~\cite{ddim} to boost the reconstruction quality and hence acquire the accurate intermediate features of subjects. We ran null-text inversion and generation for 50 timesteps. For default settings of hyperparameters, we set $\tau = 0.5$, the content generation stage is in the first 33  time steps, and then following the style generation stage. For style guidance, we adopt the same VGG-19 layers with ~\cite{adain},  and set $\lambda_s = 3.0$, $\lambda_c = 2.5$.
\subsection{Compositional Personalization Results}
\noindent\textbf{Single Subject-Style Personalization.}
As for compositional personalization, we compared our method with the latest method B-LoRA~\cite{B-lora}. We trained B-LoRA using its official code and default hyperparameters on a single image. As B-LoRA cannot be applied in a complex prompt, we only used a simple template to generate images. As shown in \Fig~\ref{fig:any_subject_any_style}, B-LoRA disrupts the structural information of the subject concept while FreeTuner achieves a harmonious blend of different concepts.

\noindent\textbf{Multiple Subject-Style Personalization.}
FreeTuner can be extended to support multiple-subject personalization. As shown in  \Fig~\ref{fig:other_uses_case}(a), different subjects can be rendered with distinct styles without affecting the background. For example, in the first row, the corgi can be rendered in style 1 and the backpack in style 2. In the second row, the styles of the subjects are interchanged. It is important to note that these images are generated using the same seed within each column.

\subsection{Single-Concept Personalization Results}

\noindent\textbf{Subject-Driven Personalization.}
 For subject-driven personalization, we compared FreeTuner with several concept customization methods, including B-LoRA~\cite{B-lora}, DreamBooth\cite{Dreambooth}, Textual Inversion~\cite{textInversion}, and SSR-Encoder~\cite{SSREncoder}. As shown in \Fig~\ref{fig:single_subject},
our training-free method is capable of faithfully capturing the details of the target concept and generating diverse images\footnote{Due to the limited space, more results are left in the Appendix. \label{ft:appendix}}. 

\noindent\textbf{Style-Driven Personalization.}
We also compared our method with recent
style transfer methods~\cite{ip-adapter, instantstyle, style_inj, StyleAligned}. As shown in \Fig~\ref{fig:style_compare}, our FreeTuner can preserve the structural information of the content image, while also transferring the style well. In contrast, other methods fail to achieve a trade-off between the transformation of style and the preservation of content. For instance, StyleAlign~\cite{StyleAligned} suffers from incorporating too many style elements and disturbs the content image's structure. StyleID~\cite{style_inj} loses the detailed information of the style image such as tones and textures due to the incorrect style image reconstruction. While IP-Adapter~\cite{ip-adapter} and InstantStyle~\cite{instantstyle} introduce the ControlNet to preserve the content image's structure, the fine-grained content details are ignored.

\subsection{Ablation Study}
\noindent\textbf{Effectiveness of Each Component.} We ablate the effectiveness of the components of FreeTuner by removing each of them. As shown in \Fig~\ref{fig:ablation_study}: 1) The intermediate features in the content generation stage are significant for preserving the content and structure of the subject. Without the features swapping, the generated corgi fails to align with the reference subject. Besides, the spatial-constrained strategy can effectively solve pixel-level artifacts, and strong visual distortion will occur without it. All these components in content generation result in high-quality subject personalization. 2) The style guidance in the style generation stage can transfer the style well and content preservation guidance can preserve the subject's visual features. Without content preservation, the generated corgi will incorporate too many style elements while ignoring the original visual appearance. 

\noindent\textbf{Generalization with Other Diffusion-Based Methods.}
Since our style guidance can be seamlessly incorporated into the denoising step without training, it can be easily combined with other diffusion-based methods to generate style-driven personalized images. \Fig~\ref{fig:other_uses_case}(b) shows examples where we combine our method with the training-based ControlNet~\cite{ControNet} and training-free BoxDiff~\cite{boxdiff}\footref{ft:appendix}. 


\section{Conclusion}
In this paper, we proposed FreeTuner, a novel, training-free approach for compositional personalization capable of generating any user-provided subject in any user-provided style. Our approach separates the generation process into two distinct stages for concept disentanglement. By injecting intermediate features to keep visual appearance of the subject and introducing style guidance to align generated images with the style concept, FreeTuner archives the preservation of both subject structure and style aesthetic features. Extensive results demonstrated FreeTuner's ability across various personalization scenarios. Moving forward, we plan to extend our framework to video generation.

\noindent\textbf{Limitations.}
While our method achieves compositional personalization in a training-free manner, there are several limitations to consider. Firstly, to acquire the accurate intermediate features, our methods adopt null-text inversion, which needs longer time than the common inversion method DDIM. Besides, due to the reliance of our personalization branch on the intermediate features of subject image reconstruction, the generation of images from multiple perspectives remains a challenging task. Finally, our style transfer capability is limited to the visual encoder.


{\small
\bibliographystyle{plain}
\bibliography{reference}
}
\clearpage
\appendix
\appendix

\section{Appendix}
In the appendix, we present additional qualitative results (Sec.~\ref{sec:quality_results}), ablation studies of other hyperparameters(Sec.~\ref{sec:ablation_study}), and more experimental details (Sec.~\ref{more_experimet})

\subsection{Additional Qualitative Results and Subject Fidelity Showcasing}
\label{sec:quality_results}
\begin{figure*}[t]
\centering
\includegraphics[width=1.0\textwidth]{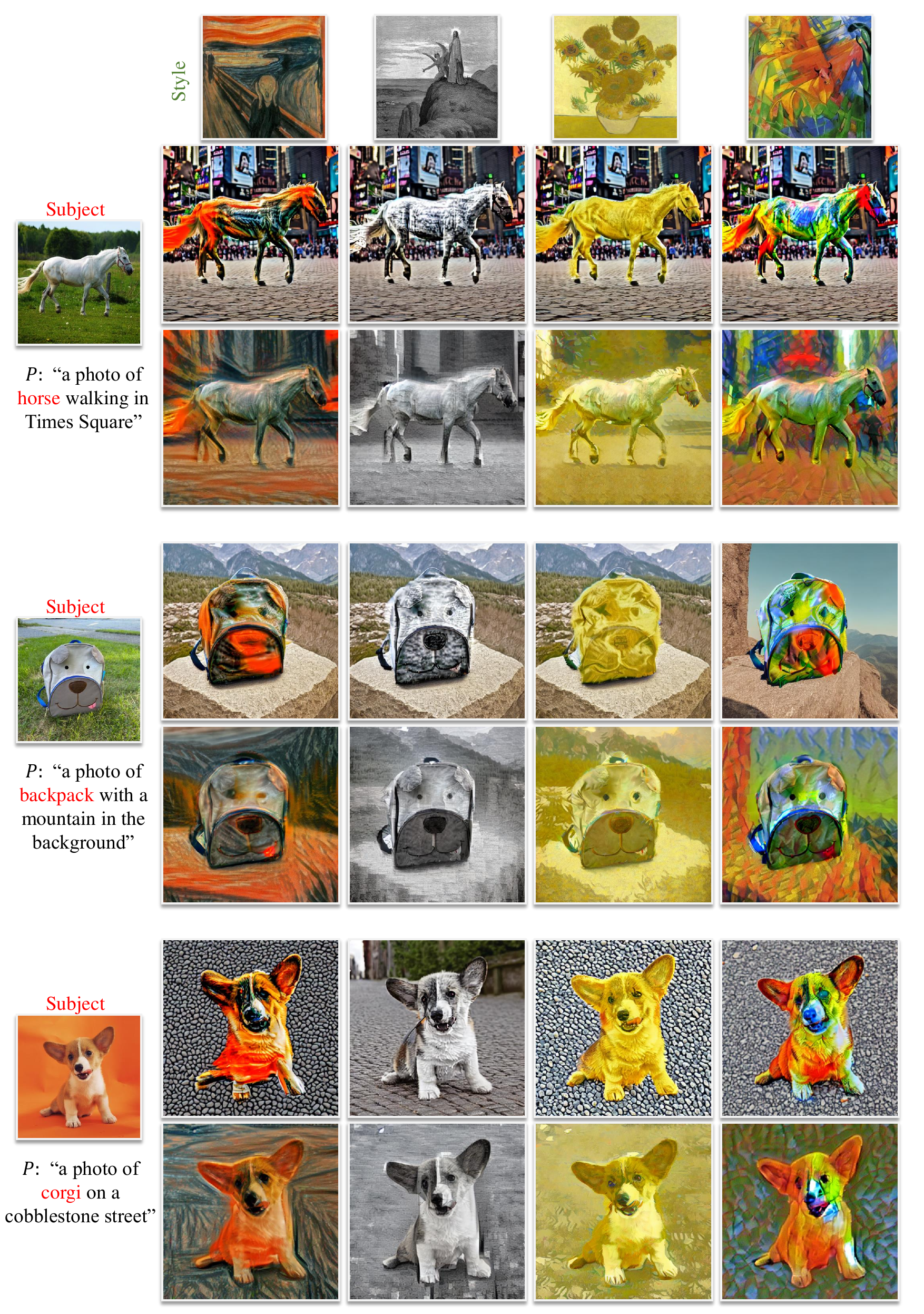}
\caption{\small  Additional qualitative results in subject-style compositional personalization. Our training-free method only needs one subject image and one style image.
}
\vspace{-0.5em}
\label{quality_compositation_personalization}
\end{figure*}

\begin{figure*}[t]
\centering
\includegraphics[width=1.0\textwidth]{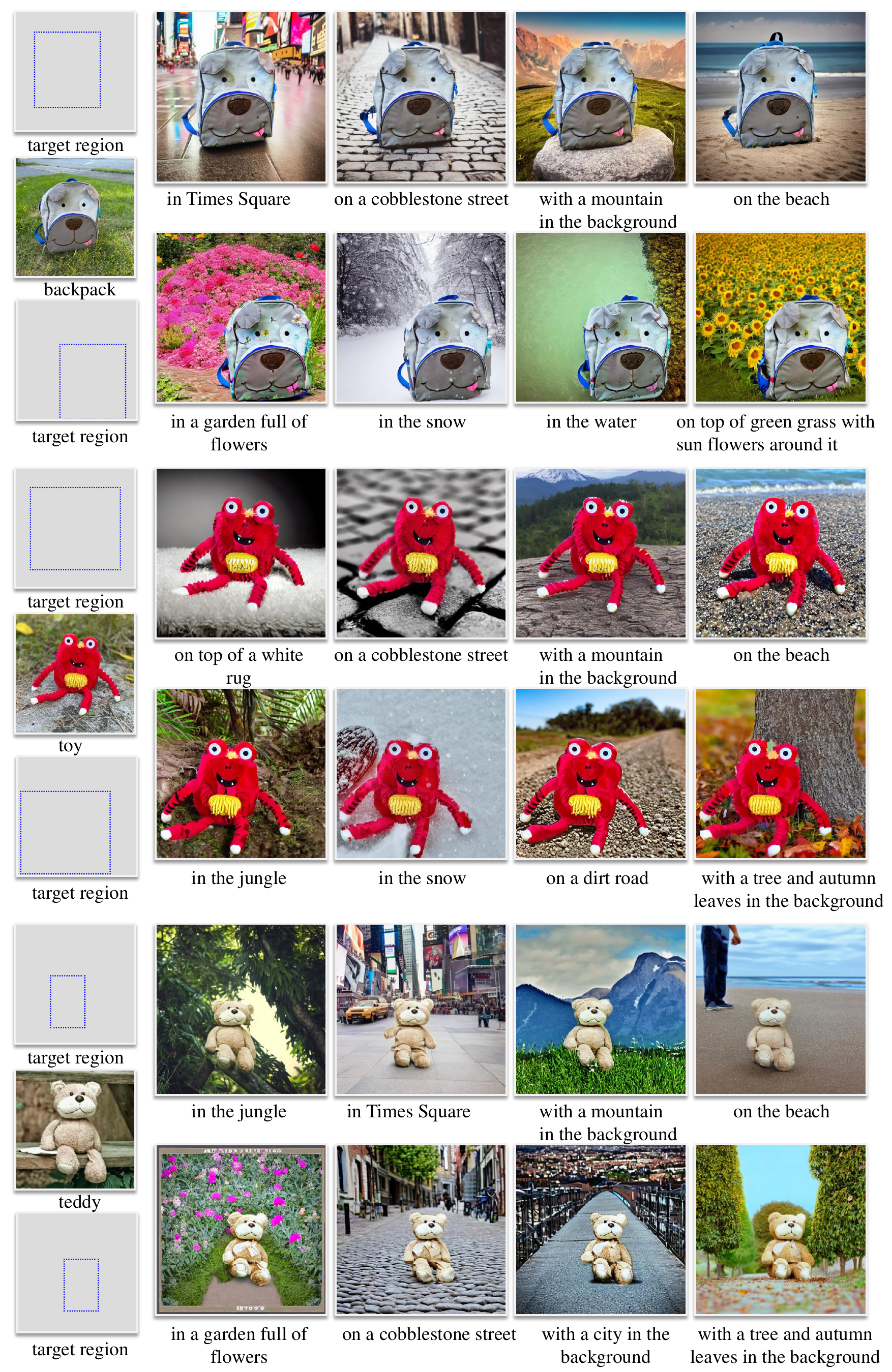}
\caption{\small  Additional qualitative results in subject-driven personalization. Our training-free method only needs one subject image for personalization and is able to control the location flexibly.
}
\vspace{-0.5em}
\label{quality_results_subject}
\end{figure*}

\begin{figure*}[t]
\centering
\includegraphics[width=1.0\textwidth]{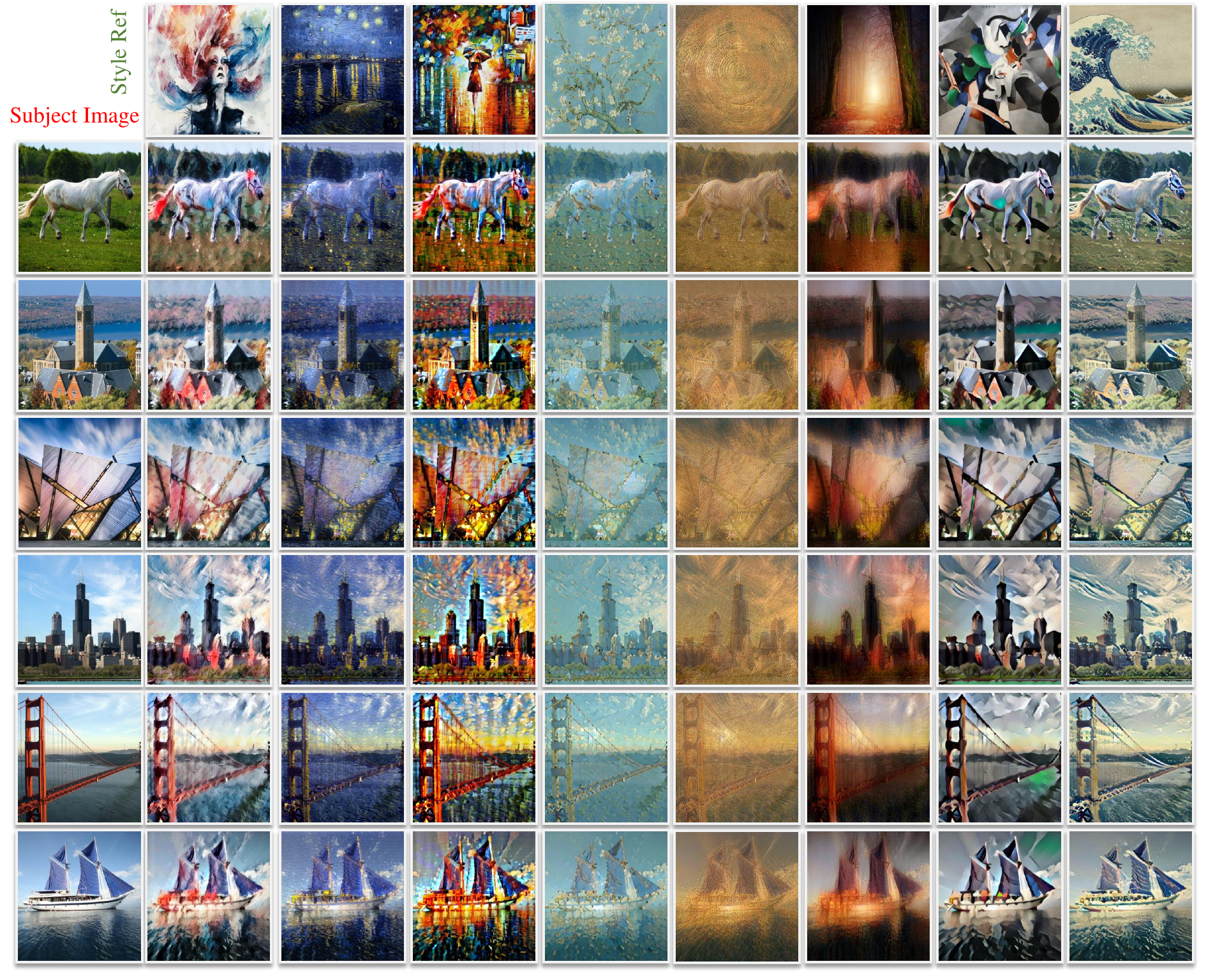}
\caption{\small  Additional qualitative results in style-driven personalization. Zoom in for viewing details.
}
\vspace{-0.5em}
\label{quality_results_style}
\end{figure*}

Our FreeTuner focuses on three main applications: subject-driven personalization, style-driven personalization, and subject-style compositional personalization.
In \Fig~\ref{quality_compositation_personalization}, we present additional results generated for compositional personalization based on subject and style image. The first column is the subjects, while the first row corresponds to the style image. Our method can render the entire image or just the subject within the generated image.
In \Fig~\ref{quality_results_subject}, we provide more subject-driven 
personalization results, the first column is the subject image and the user-provided target region, and the others are the personalized images. Our training-free method can generate the subject in the target region while aligning well with the prompt.
In 
\Fig~\ref{quality_results_style}, we provide additional qualitative results in style-driven personalization. We provide the style concept in the first row and the content image in the first column, in the rest columns, we provide transferred images.

\subsection{Additional Ablation Study}
\label{sec:ablation_study}
\begin{figure*}[h]
\centering
\includegraphics[width=1.0\textwidth]{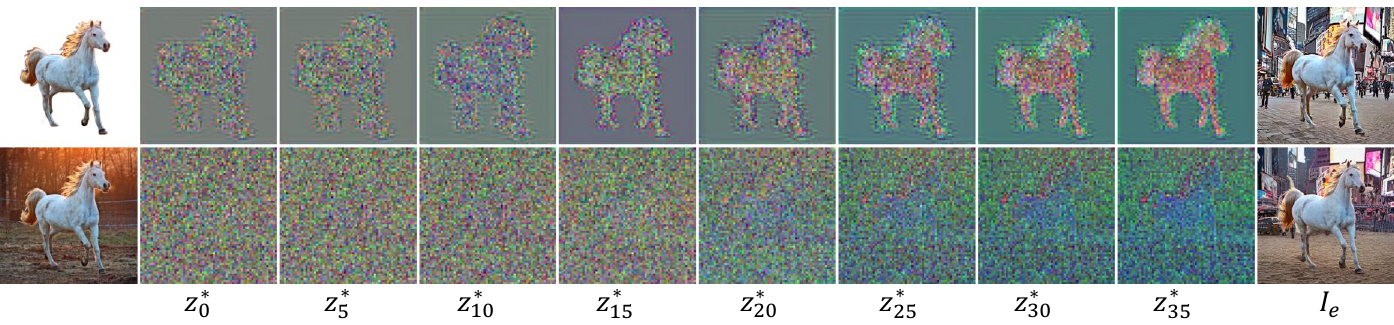}
\caption{\small The influence of subject-preprocess on the latent code and the final generated personalized image.
}

\vspace{-0.5em}
\label{effecct_of_process_to_latent}
\end{figure*}

\begin{figure*}[h]
\centering
\includegraphics[width=1.0\textwidth]{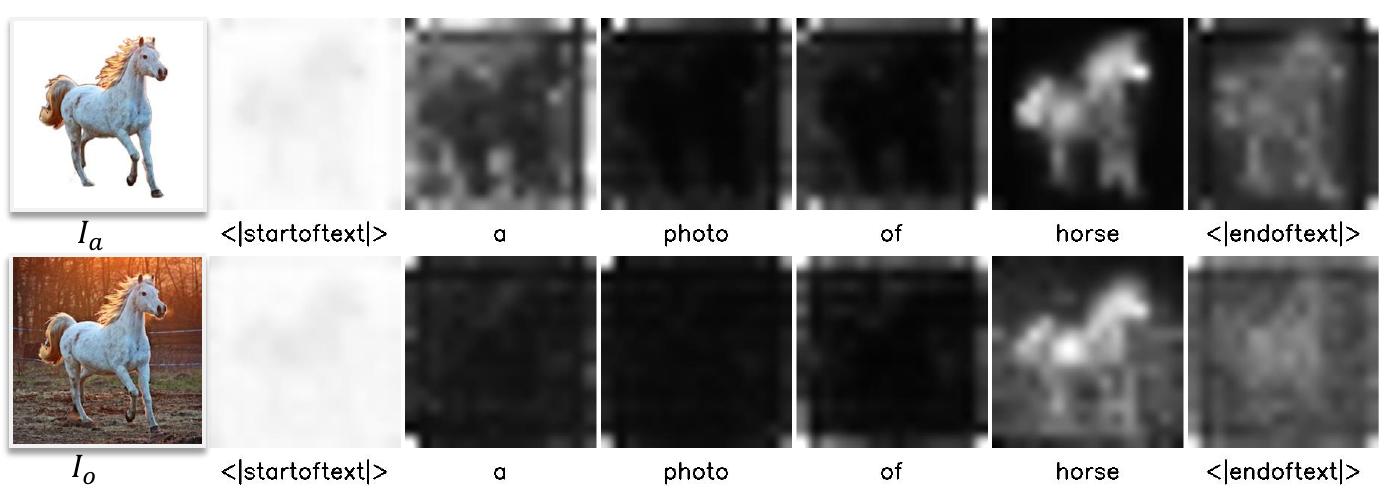}
\caption{\small The influence of subject-preprocess on the cross-attention maps.
}
\vspace{-0.5em}
\label{effect_of_preprocess_to_cross}
\end{figure*}

\noindent\textbf{Effect of Subject-Preprocess.} \Fig~\ref{effect_of_preprocess_to_cross} shows the influence of subject-preprocess on cross-attention maps, it is difficult to distinguish the subject and background without the subject-preprocessing operation. \Fig~\ref{effecct_of_process_to_latent} shows the influence of subject-preprocessing on latent code $z$ and generated image.  Without the subject-preprocessing operation, the background of the original image will influence the generated personalized image.

\begin{figure*}[h]
\centering
\includegraphics[width=1.0\textwidth]{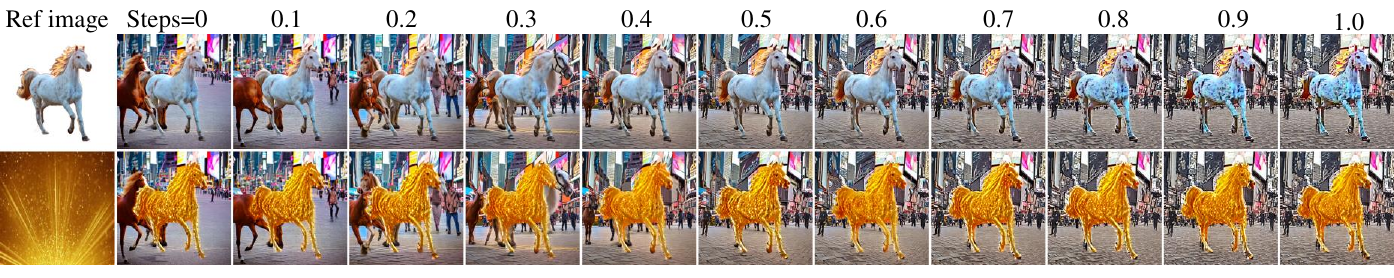}
\caption{\small  Number of attention features injection steps $\tau$.
}
\vspace{-0.5em}
\label{effecct_of_attention_ing_to_generated}
\end{figure*}

\noindent\textbf{Number of Attention Features Injection Steps $\tau$.} \Fig~\ref{effecct_of_attention_ing_to_generated} demonstrates the number of attention features injection steps related to the content and layout of the generated image. The presence of pixel-level artifacts is linked to a low number of injection steps, while an increase in the number of steps may result in visual distortion. To get a balance, we set $\tau = 0.5$.

\subsection{More Experimental Details}
\label{more_experimet}
\begin{figure*}[t]
\centering
\includegraphics[width=1.0\textwidth]{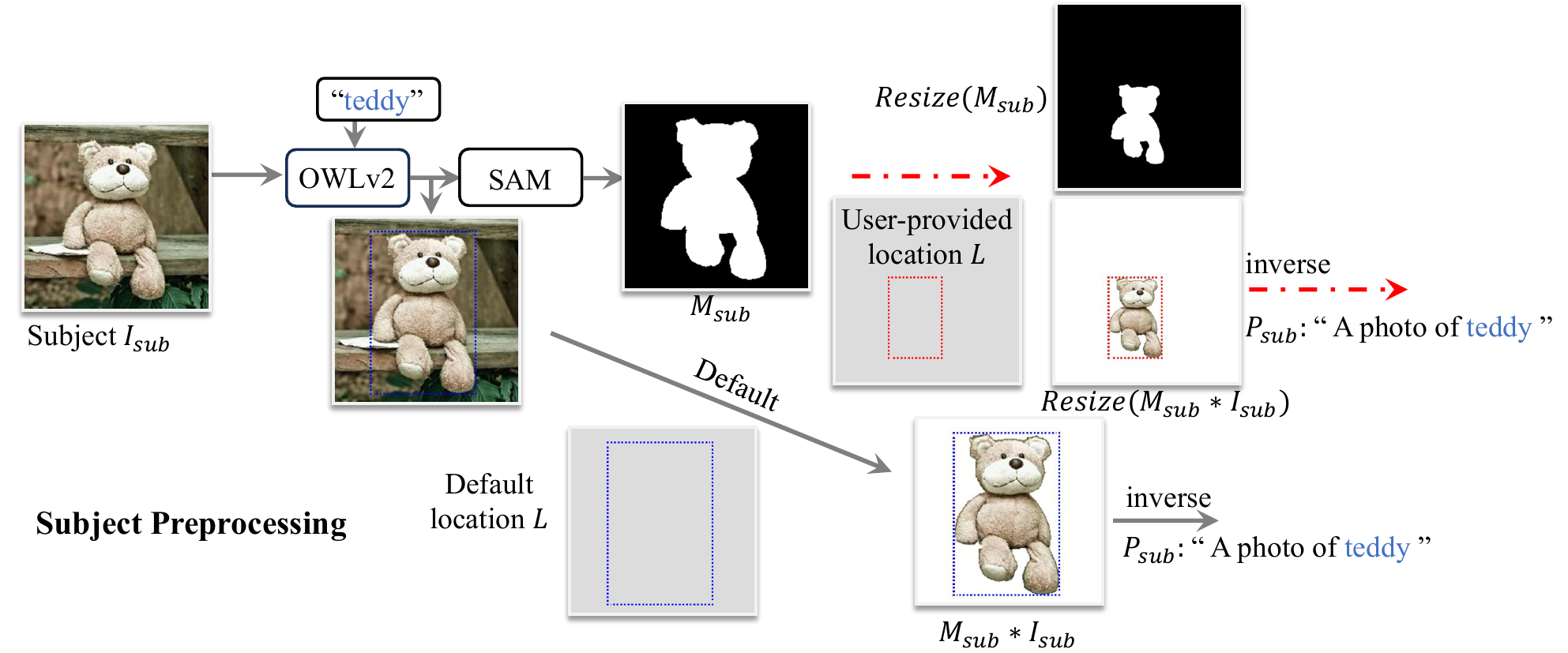}
\caption{\small Subject preprocessing operation.}
\vspace{-0.5em}
\label{preprocess}
\end{figure*}

\noindent\textbf{Subject Preprocessing.} 
We adopt the automatical pipe presented by MuDI~\cite{mudi} to extract the segmentation map of the user-provided subject. Specifically, this method begins with the extraction of subject bounding boxes using the OWLv2~\cite{owlv2}
, Subsequently, SAM~\cite{sam} segments the subjects based on these bounding boxes. 
\Fig~\ref{preprocess} shows the detail of the preprocessing. After getting the binary mask, we can remove the background of the subject image directly or resize the mask and subject in the user-provided location $l$.  Then we inverse the image with a simple prompt containing the subject's class name.

\noindent\textbf{Prompts Used in the Experiments.} 
For the subject-driven qualitative evaluation as shown in  \Fig~\ref{fig:single_subject}, we adopt the following text prompts. For the subject \texttt{``corgi"}, we use \texttt{``a photo of corgi in Times Square", ``a photo of corgi near the lake", ``a photo of corgi in the Acropolis".} For the subject \texttt{``toy"}, we use \texttt{``a photo of toy in a garden full of flowers", ``a photo of toy on a cobblestone street", ``a photo of toy in the jungle"}.

\noindent\textbf{Spatial-constrained Strategy.} 
The energy function in Spatial-constrained Strategy is built on the methodology presented by BoxDiff~\cite{boxdiff}, which proposes Inner-Box, Outer-Box, and
Corner Constraints on cross-attention maps to achieve e a training-free layout-to-image generation.
\noindent\emph{\underline{1) Inner-Box Constraint:}}
To ensure the synthesized objects will approach the user-provided locations. BoxDiff proposes the inner-box constraint:
\begin{equation}\mathcal{L}_{IB}=\sum_{w_i \in P_{comp}}\left(1-\frac{1}{S}\sum\mathbf{topk} (\widetilde{CA}_{t}^i\cdot\mathbf{M}_i,S)\right),
\label{Inner-Box}
\end{equation}
where $\mathbf{topk}(\cdot, S)$ means that $S$ elements with the highest response in input would be selected and $M_i$ is the user-provided region for the subject.

\noindent\emph{\underline{2) Outer-Box Constraint:}} To prevent the object from moving out of the target regions, BoxDiff proposes the outer-box constraint:
\begin{equation}\mathcal{L}_{OB}=\sum_{w_i \in P_{comp}} \left(1-\frac{1}{S}\sum\mathbf{topk}(\widetilde{CA}_{t}^i\cdot(1-\mathbf{M}_i),S)\right),
\label{Outer-Box}
\end{equation}

\noindent\emph{\underline{3) Corner Constraint:}}
Moreover, to ensure the objects fill the entire box.
BoxDiff proposes the corner constraint 
$\mathcal{L}_{cc}$ at the projection of the $x$-axis and $y$-axis. $\mathcal{L}_{cc}$ computes the projection difference between the target mask $M_i$ and cross-attention map $\widetilde{CA}_{t}^i$.

\begin{figure*}[h]
\centering
\includegraphics[width=1.0\textwidth]{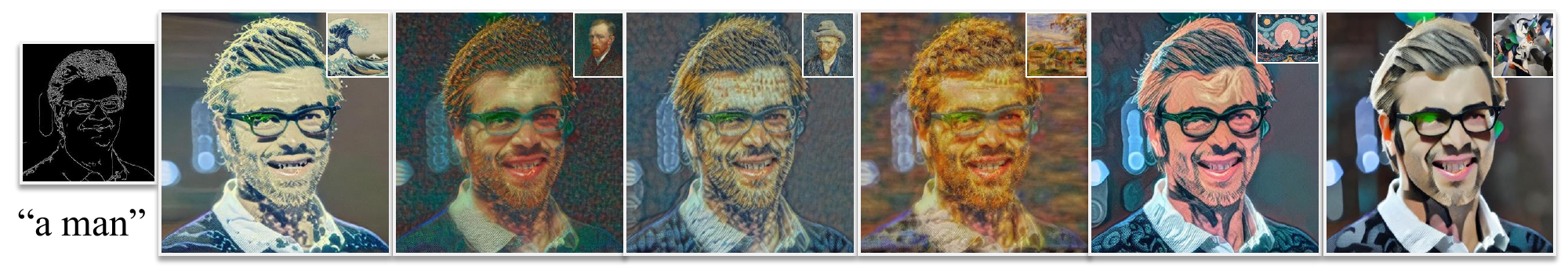}
\caption{\small  Additional qualitative results.
}
\vspace{-0.5em}
\label{quality_results_ControlNet}
\end{figure*}

\end{document}